\documentclass[lettersize,journal]{IEEEtran}
\usepackage{amsmath,amsfonts}
\usepackage{algorithmic}
\usepackage{algorithm}
\usepackage{array}
\usepackage[caption=false,font=normalsize,labelfont=sf,textfont=sf]{subfig}
\usepackage{textcomp}
\usepackage{stfloats}
\usepackage{url}
\usepackage{verbatim}
\usepackage{graphicx}
\usepackage{cite}
\usepackage{float}
\usepackage{multirow}
\usepackage{booktabs}
\usepackage[utf8]{inputenc}
\usepackage{amsmath}
\usepackage{diagbox}
\usepackage{hyperref}

\begin{document}

\title{Cross-Modality Feature Fusion Based on Structured State Space Duality for Multimodal Image Registration Network}

\author{Zhikang Li, Yan Wu, Xin Hu, Yi Dai, Ming Li        
\thanks{This work was supported in part by the Natural Science Foundation of China under Grant 62172321. (Corresponding author: Yan Wu.)

Zhikang Li, Yan Wu, Xin Hu and Yi Dai are with the Remote Sensing Image Processing and Fusion Group, School of Electronic Engineering, Xidian University, Xi’an 710071, China. (e-mail: ywu@mail.xidian.edu.cn)

Ming Li is with the National Key Laboratory of Radar Signal Processing, Xidian University, Xi’an 710071, China.}
}



\maketitle

\begin{abstract}
In multi-modal image registration, the primary challenge lies in shared structural information extraction. Compared to Transformers, Structured State Space Duality (SSD) offers greater global structural feature extraction with higher efficiency during training and inference. Inspired by these advantages, we propose a novel algorithm for multi-modal image registration, named RegNetMamba-2. Our algorithm incorporates SSD into coarse-to-fine matching process to extract local and global structural features effectively. Firstly, SSD is applied in three different scales for multi-modal feature extraction in our network. To strengthen local representation, we pay more attention on foreground edge and structural information by feature scaling function of SSD. Secondly, for shared feature extraction of input images and multi-modal feature fusion in all scales, we propose cross-modality feature fusion model based on SSD, consisting of Cross-Modality feature Interaction (CMI) module and Multi-Scale feature Fusion (MSF) module. CMI module is designed for cross-modality feature extraction of each scale by SSD in cross form. MSF module is designed to employ a progressive upward fusion in feature-level to obtain fine features, consisting of multi-modal features in all scales. Following coarse-to-fine, the features in 1/8 scale from CMI and 1/2 scale from MSF are collected to calculate matching probability scores. Then we respectively establish matching process by correspondences of pixel-wise. Extensive experiments demonstrate that comparing with state-of-the-art deep-learning based algorithms, RegNetMamba-2 has achieved good effects in both performance and efficiency for multi-modal image registration on the following datasets: VIS-SAR (OSDataset), VIS-IR (LGHD/RoadSence) and VIS-NIR (RGB-NIR sense). 
\end{abstract}


\section{Introduction}
\IEEEPARstart{I}{mage} registration serves as a critical prerequisite for a wide range of computer vision tasks, particularly in Structure-from-Motion (SfM) \cite{ref1} and Simultaneous Localization and Mapping (SLAM) \cite{ref2}. In recent years, multi-modal image processing has garnered growing interest owing to its ability to provide complementary information acquired from diverse sensors. Multi-modal image registration forms the foundational step for subsequent applications such as image fusion \cite{ref3}, image classification \cite{ref4}, and change detection \cite{ref5}. Nevertheless, this process is challenged by inherent geometric variations, including rotation, translation and scaling, as well as significant modal discrepancies and sensor-specific noise. For instance, synthetic aperture radar (SAR) sensors offer all-weather and multi-temporal capabilities, yet SAR imagery is often degraded by multiplicative speckle noise. Long-wave infrared (IR) sensors enable operation around the clock independent of lighting conditions, but IR images typically exhibit blurred edges and are susceptible to additive noise. Near-infrared (NIR) imagery, while providing higher resolution and relatively lower noise levels than IR, still manifests substantial modality gaps when compared with visible images.

Depending on whether a detector is employed in the matching pipeline, image registration methods can be broadly categorized into sparse or semi-dense matching approaches. Detector-based methods constitute sparse matching strategies, which typically adhere to a detect-then-describe or a joint detection and description paradigm. Most handcrafted methods first detect keypoints in the input images and then compute descriptors based on gradient or phase congruency information, such as GLF-MIFT \cite{ref6}, POS-GIFT \cite{ref7}, AMES \cite{ref8}, RIFT \cite{ref9}, and CoFSM \cite{ref10}. Similarly, patch-based deep-learning methods also follow the detect-then-describe workflow, but leverage Convolutional Neural Networks (CNNs) to extract descriptors from image patches, as seen in MRAM \cite{ref11}, Cnet \cite{ref12}, EFRNet \cite{ref13}, Rotation Invariant Descriptors \cite{ref14}, and SSML-QNet \cite{ref15}. Another pipeline of deep-learning based registration methods adopts the joint detection and description framework, where a siamese CNN encoder simultaneously predicts keypoint locations and their descriptors; representative works include SuperPoint \cite{ref16}, D2-Net \cite{ref17}, R2D2 \cite{ref18}, and ReDFeat \cite{ref19}. A fundamental requirement for multi-modal image registration is the extraction of shared features across modalities. However, the CNN-based methods mentioned above are inherently limited by local receptive fields and a lack of explicit cross-modality interaction, thus constraining their ability to capture modality-invariant features. Moreover, the descriptors obtained from CNNs often suffer from low discrimination and are difficult to optimize, thus adversely affecting feature matching.

To address this issue, graph matching methods have been proposed to enhance feature distinctiveness by learning correspondences through Graph Neural Networks (GNNs) or Transformers. Notable examples include SuperGlue \cite{ref20}, LightGlue \cite{ref21}, OmniGlue \cite{ref22}, ClusterGNN \cite{ref23}, SeedGNN \cite{ref24}, and FeatureBooster \cite{ref25}. These GNN-based methods significantly improve the matching accuracy of sparse descriptors by leveraging global aggregation and cross-feature interactions during self-attention and cross-attention. However, they still rely on high-quality keypoints and descriptors as inputs.

Conversely, semi-dense matching methods eliminate the need for explicit feature detection and instead apply Transformer architectures directly over entire feature maps to establish global interactions between input images. Representative approaches include LoFTR \cite{ref26}, MatchFormer \cite{ref27}, ASpanFormer \cite{ref28}, ASTR \cite{ref29}, XoFTR \cite{ref30}, LoFLAT\cite{ref47} and JamMa\cite{ref45}. Most of these methods follow a coarse-to-fine pipeline: (1) local feature extraction using CNNs, (2) coarse-level interaction and matching via a Transformer-based network in 1/8 resolution, (3) fine-level refinement and matching in 1/2 resolution using window-based attention. By involving all features in interaction and matching, these algorithms enable the estimation of dense correspondences. However, the multi-head attention in standard Transformers exhibits high computational complexity and limited robustness to noise, while linear attention suffers from smooth distribution of attention, which hinders their effectiveness in multi-modal image matching. To deal with this problem, \cite{ref46} proposes Focus Linear Attention to improve discriminability and diversity of features for local representation enhancement. LoFLAT\cite{ref47} introduces the attention of \cite{ref46} into LoFTR architecture to improve the performance of semi-dense local matching while preserving low computational complexity.

Recently, the Mamba architecture based on Structured State Space Models (SSMs) \cite{ref31} has emerged as a promising approach. It achieves global receptive fields with linear computational complexity through a selective scanning mechanism and has been widely adopted in computer vision tasks such as VMamba \cite{ref32} LocalMamba \cite{ref33} and MSVMamba \cite{ref34}. Thanks to its forget gate and SSM block design, Mamba demonstrates superior performance compared to linear attention \cite{ref35}. However SSMs often train less efficiently than Transformers. To address this issue \cite{ref36} proposed Structured State Space Duality (SSD) which connects structured SSMs with attention variants. Building on SSD, the Mamba-2 architecture was developed and outperforms original Mamba in both performance and efficiency. Furthermore \cite{ref37} adapted SSD for visual applications by transforming it into a non-causal form suitable for image data.

Multi-modal image registration is a computationally intensive process and has to cope with significant modality differences, particularly in SAR and IR. Currently, image matching methods typically require large-scale datasets, so enhancing training efficiency and reducing computational costs are the key issues to be addressed. SSD connected SSMs and attention, which strengthens the global perspective while maintaining relatively lower complexity. Compared to Transformers, SSD enables more efficient training and inference. Inspired by these advantages, we introduce SSD into the coarse-to-fine detector-free matching process for the first time and propose our novel algorithm named RegNetMamba-2. Vanilla SSD can provide superior global structural feature extraction but local representation is insufficient, which will sometimes lead to over-smoothing. To solve this problem, we pay more attention on foreground edge and structural information by feature scaling function of SSD, allowing us to improve feature similarity distribution for local feature enhancement. Before the process of scanning, input tokens are scaled and re-normalized, which pushes apart irrelevant features while pulling closer correlated ones. 

In RegNetMamba-2, we proposed a cross-modality feature fusion model to extract coarse and fine features, which contains two modules: Cross-Modality feature Interaction
(CMI) module and Multi-Scale feature Fusion (MSF) module. CMI module is designed for shared structural feature extraction through SSD. Previous semi-dense methods most depend on Transformer with self and cross-attention, while vanilla Mamba-2 lacks a cross mechanism. According to the relationship between SSMs and attention, we extend SSD into cross form to enable cross-modality interaction. In CMI module, we adopt SSD in each scale to extract cross-modal features and fuse them into multi-modal features. In MSF module, features in three different scales extracted by CMI module will be fused progressively upwards by SSD to get fine features in feature-level. Following coarse-to-fine, to respectively establish matching process by correspondences of pixel-wise, the features in 1/8 scale from CMI module and 1/2 scale from MSF module are collected to calculate matching probability scores.

Our contributions can be summarized as follows:

(1). We propose a novel detector-free method named RegNetMamba-2. In our network, SSD is incorporated into coarse-to-fine matching process for multi-modal image registration to extract local and global structural feature with more efficient training and inference.  

(2). For local enhancement, we pay more attention on edge and structural information by feature scaling function of SSD. This function improves the distribution of attention in SSD, enhancing discriminability of local features.

(3). A novel cross-modality feature fusion model is constructed, consisting of CMI and MSF modules. CMI module is designed for shared structural feature extraction, which extends SSD into cross form and applies to the feature maps in each scale. MSF module aims to fuse multi-modal features extracted by CMI in three different scales through SSD.

\section{relative works}
\subsection{Detector-based image registration methods}
The typical detector-based image registration pipeline is as follows: keypoints detection, descriptor extraction, feature matching, and transform estimation. RIFT  \cite{ref9} detects keypoints on maximum moment map by FAST \cite{ref38} and extract descriptors in Maximum Index Map (MIM) of 6 orientations to achieve radiation and rotation insensitive, while the whole process is based on Log-Gabor filter and phase congruency. With the advancement of deep learning, detectors and descriptors have gradually been replaced by neural networks. Cnet \cite{ref12} introduces channel and spatial attention into CNNs and proposes a correlation loss to extract the invariant features of multi-modal image patches.

The powerful learning capabilities of neural networks have introduced a new paradigm for sparse matching: joint detection and description. A prominent example is D2-Net \cite{ref17}, which employs a shared CNN backbone as the encoder. Each feature vector serves as a descriptor, while detector scores are derived directly from the descriptor feature maps. R2D2 \cite{ref18} further enhances the reliability and repeatability of D2-Net. Building on this, ReDFeat \cite{ref19} has recoupled the detection and description processes through mutual weighting based on the R2D2 framework.

To further improve feature representation and discriminative, graph matching methods based on GNNs or Transformers have been proposed. SuperGlue \cite{ref20} utilizes graph attention networks to aggregate global features and model interactions among input features, while employing the Sinkhorn algorithm to estimate correspondences from similarity scores. To enhance the accuracy and efficiency of graph matching, LightGlue \cite{ref21} introduces an early-exit operation and prunes low-confidence points to speed up inference. FeatureBooster \cite{ref25} proposes a novel attention-free transformer architecture that achieves linear complexity for global feature aggregation. Nevertheless, graph matching approaches are heavily dependent on the availability of high-quality keypoints and descriptors extracted beforehand, posing a significant challenge in multi-modal image registration.

\subsection{Detector-free image registration methods}
Unlike detector-based pipelines, detector-free matching methods entirely bypass the explicit extraction of keypoints. By incorporating all features into the matching process, they are capable of producing dense pixel-wise correspondences. LoFTR \cite{ref26} is the first to introduce Transformers into semi-dense matching, adopting a coarse-to-fine approach that computes initial matches at 1/8 scale and refines them within local windows on 1/2 scale feature maps. Building on this architecture, ASpanFormer \cite{ref28} introduces local-global attention and local cross-attention with adaptive spans at the coarse level; ASTR \cite{ref29} proposes spot-guided attention based on local matching consistency; XoFTR \cite{ref30} designs a novel fine-level matching module using window cross-attention to fuse features from different scales to enhance fine-grained matching.

\begin{figure*}[t]
\centering
\includegraphics[width=\textwidth]{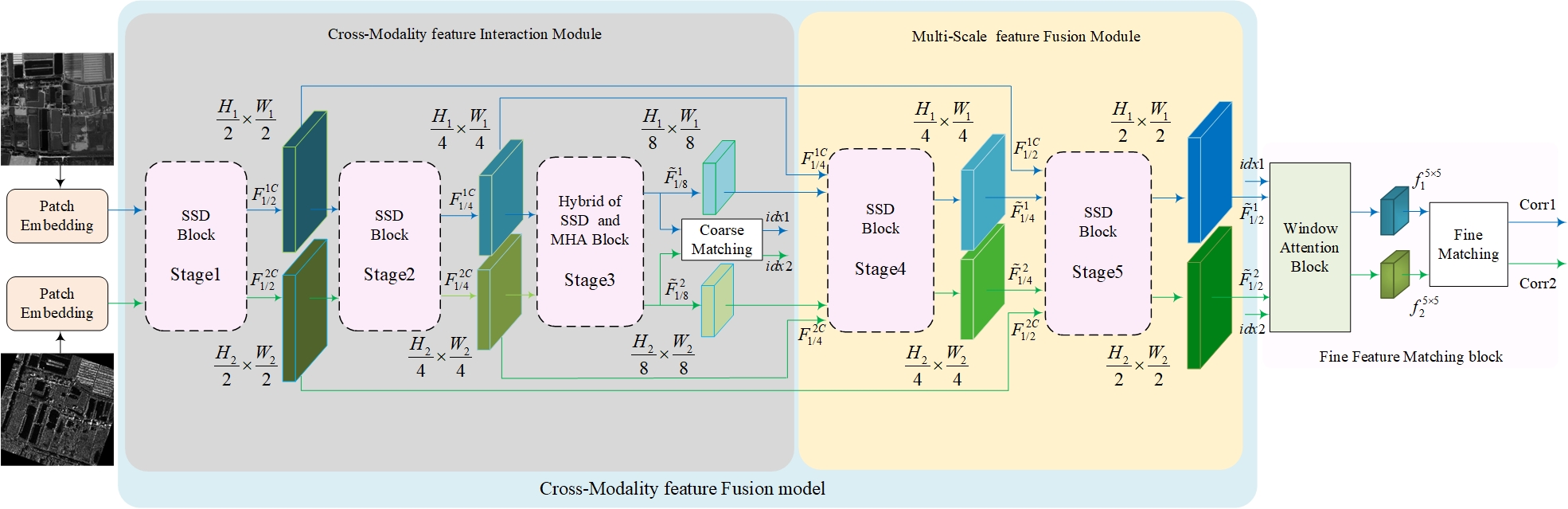}
\caption{Overall architecture of our RegNetMamba-2. CMI modules contains Stage1 to Stage3 for feature extraction in scale of 1/2, 1/4 and 1/8. After Stage3 we gain the coarse features for coarse matching in scale of 1/8. MSF module contains Stage4 and Stage5 for feature fusion in different scales and get the features in 1/2 scale. After Stage5 we apply the last window attention and calculate fine correspondences.}
\label{Fig.1}
\end{figure*}

State Space Models (SSMs) are increasingly being adopted in vision tasks due to their ability to capture global features with linear complexity. Recently, SSMs have also been explored for cross-image feature interaction in matching tasks. For instance, JamMa \cite{ref45} proposed a Joint Mamba approach that concatenates two inputs both horizontally and vertically and performs integrated scanning in four directions. While this scanning strategy enables mutual interaction, it inevitably breaks the global information of the images and may introduce irrelevant context. Image registration requires the extraction of shared structural features from image pairs. However, this goal is difficult to achieve with the selective scanning mechanism of SSMs, which struggles to effectively facilitate interaction between two inputs. Furthermore, image registration often demands large datasets for training. As mentioned in \cite{ref36}, compared to Transformers, SSMs can be more challenging to be trained efficiently, often incurring higher training costs.

Therefore, Structured State Space Duality (SSD) was proposed to formalize the connection of SSMs and variants of attention. SSD offers a superior global receptive field and significantly improves training and inference efficiency through structured matrix algorithms. Inspired by these advances, we incorporate SSD into detector-free semi-dense matching pipeline and propose a novel algorithm named RegNetMamba-2 for multi-modal image registration to extract local and global structural features effectively.

\section{Methodology}
In this section, we will introduce our novel algorithm  RegNetMamba-2. Firstly, we introduce the overview of our network. Our feature extraction model is called cross-modality feature fusion model,  CMI and MSF module. Then, we explain the local enhancement and cross form of SSD, which is the basic block of our network. After that, we display the process of feature extraction by CMI, as well as the multi-scale feature fusion by MSF. Finally, we briefly introduce coarse-to-fine feature matching process.

\subsection{RegNetMamba-2 Architecture}
RegNetMamba-2 is a detector-free matching algorithm following a coarse-to-fine process, as shown in Fig.\ref{Fig.1}. In our method, cross-modality feature fusion model is designed for cross-modality feature extraction and multi-modal feature fusion in different scales, containing CMI module and MSF module. Previously, CNN patch embedding block projects the images into initial feature maps in the scale of 1/2. Firstly, CMI module is designed to extract shared structural features by cross-modal interaction based on SSD. In CMI module, SSD is extended into cross form and applied in different scales from Stage1 to Stage3, which allows us to extract cross-modality features and fuse them into multi-modal features in each scale. Secondly, after CMI module, MSF module is designed to progressively fuse multi-modal features upwards in three scales. Similar to CMI, features in smaller scale will be fused into larger scale by multi-scale feature interaction based on SSD from Stage4 and Stage5. After MSF module, we get fine features in the scale of 1/2. Following coarse-to-fine process, coarse matching matrix is estimated based on feature maps in 1/8 scale extracted by CMI module, then fine matching probability scores will be calculated from features in 1/2 scale gained by MSF module. Finally, we respectively establish matching process by correspondences of pixel-wise.

\subsection{Cross-modality feature fusion model}
SSD has a superior global representation with linear complexity, since the structured matrix algorithms has connected SSMs and attention. Based on this research, we propose cross-modality feature fusion model depending on SSD. Previously, we introduce the basic block of our network: local enhanced SSD. After that, we display two core elements of our cross-modality feature fusion model: CMI module and MSF module.

\subsubsection{Local Enhanced SSD} \label{sec:block}
Firstly, we review the principles of SSM and SSD. The vanilla SSMs \cite{ref31} can be equalized as follows:
\begin{align}\label{Eq 1}
h(t) & = Ah(t-1)+Bx(t),\nonumber\\
y(t) & = Ch(t)
\end{align}

SSD \cite{ref36} has quadratic mode by vectorizing M, then Eq. (\ref{Eq 1}) can be represented as:
\begin{align}\label{Eq 2}
Y=(L\circ CB^{T} ) X
\end{align}

Unlike NLP, casual order based on scanning paths will break global context of an image. In \cite{ref37}, vanilla SSD is turned into non-casual form, where comulative multiplication of A is eliminated while forward and backward scanning paths are combined as:
\begin{align}\label{Eq 3}
H _{i}=\sum_{j=1,j\ne i}^{N} \frac{1}{A_{j}} B_{j}x(j) + \frac{1}{A_{i}} B_{i}x(i)
\end{align}

Finally, matrix multiplication of SSD in vision is:
\begin{align}\label{Eq 4}
Y=CH=C(B^{T}(X\cdot m) ,m=\frac{1}{A}    
\end{align}

However, SSD suffers from low discriminability of features similar to linear attention and lacks local representation, which will lead to severe over-smoothing problem, especially for night optical images and IR images. To deal with this problem, we lead SSD to pay more attention to edge and structural features by feature scaling function for local enhancement. The similarity of token B and token C is defined as follows:
\begin{align}\label{Eq 5}
sim_{s}(C,B)=f_{s}(C)f_{s} (B)^{T} 
\end{align}

if $x = (x_{1},x_{2},...,x_{d})$, then $f_{s}(x)$ is calculated as:
\begin{align}\label{Eq 6}
f_{s} (x)=x^{p} \cdot \sqrt{\frac{\sum_{i=1}^{d}x_{i}^{2}  }{\sum_{i=1}^{d}x_{i}^{2p}} }
\end{align}

Parameter \textit{p} control the degree of feature scaling. \textit{p} is manually set satisfying $\textit{p}>1$. When $C_{i}$ and $B_{j} $ are the corresponding tokens, $sim_{s} (C_{i} ,B_{j} )> C_{i} B_{j}^{T}$, while if $C_{i}$ and $B_{j} $ have low correlation, there satisfies $sim_{s} (C_{i} ,B_{j} )< C_{i} B_{j}^{T}$.

For the proof of above conclusion, we can assert that C and B are normalized, which satisfies $\left \| C \right \| _{F}$=1,  $\left \| B \right \| _{F}$=1. We divide both the numerator and the denominator by $c_{max}$ and $b_{max}$, where $c_{max}=max(c_{i})$, $b_{max}=max(b_{i})$, then:

\begin{align}\label{Eq 7}
&\lim_{p \to \infty}(sim_{s} (C,B))= \lim_{p \to \infty}\frac{C^{p} }{\sqrt{\sum_{i = 1}^{d} c_{i}^{2p}  } } \cdot \frac{B^{p} }{\sqrt{\sum_{i = 1}^{d} b_{i}^{2p}  } }&\nonumber \\ 
&= \lim_{p \to \infty}\frac{ {\textstyle \sum_{i = 1}^{d}} ({c_{i}/c_{max}) ^{p} \cdot (b_{i}/b_{max})^{p} }}{\sqrt{\sum_{i = 1}^{d} (c_{i}/c_{max})^{2p}}\sqrt{\sum_{i = 1}^{d} (b_{i}/b_{max})^{2p}}} &\nonumber \\  
&= \lim_{p \to \infty}\frac{ {\textstyle \sum_{i = 1}^{d}} ({u_{i} v_{i})^{p} }}{\sqrt{\sum_{i = 1}^{d} u_{i}^{2p}}\sqrt{\sum_{i = 1}^{d} v_{i}^{2p}}}&
\end{align}

when C and B are highly correlative, there exists $m$ satisfying $u_{m} = v_{m} =1$, then we can get:
\begin{align}\label{Eq 8}
\lim_{p \to \infty} sim_{s} (C,B) =1> CB^{T} 
\end{align}

if B and C are irrelevant and \textit{m} is not exist, obviously:
\begin{align}\label{Eq 9}
\lim_{p \to \infty}(u_{i} v_{i})^{p} =0, \lim_{p \to \infty} sim_{s} (C,B)=0 < CB^{T}
\end{align}

Therefore, if we set an appropriate $\textit{p}>1$, our scaling function will improve the relevance distribution of SSD, which promotes the similarity of correlated features and pull away irrelevant features. During global feature interaction, this function can lead SSD to concentrate more on foreground edge and structural information, thus improving the representation of local features. In our method, \textit{p} is set to 2.

Vanilla SSD has no cross form like cross-attention. To enable interaction of cross-modality and fusion of multi-scale features without breaking global context, we extend SSD into cross form according to the relationship between SSMs and attention. Our SSD in cross form can be formulated as follows:
\begin{align}\label{Eq 10}
& H_{2} = \sum_{j = 1}^{L} \frac{1}{A_{j} } B_{j} X_{2} (j)\nonumber \\
& Y_{1} = C_{1} H_{2}+DX_{1} 
\end{align}

\begin{figure}[t] 
\centering
\includegraphics[width=0.45\textwidth]{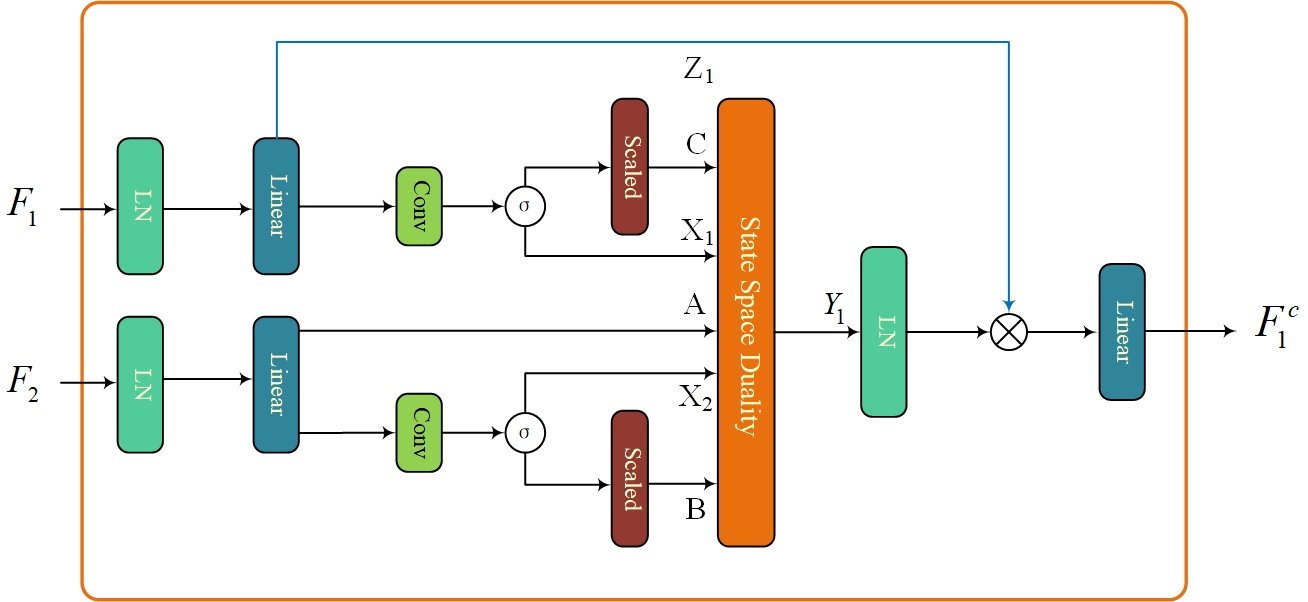}
\caption{Architecture of our local enhanced SSD layer in cross form.}
\label{Fig.2}
\end{figure}

We set $G(x) = GeLU(Conv(Linear(x))$. The group of \textit{A}, \textit{B}, $X_{2} $ come from target input $F_{2} $, which can be calculated as follows:
\begin{gather} \label{Eq 11}
A=softplus(Linear(F_{2} ))\nonumber\\
X_{2} =G_{2}(F_{2}), B=f_{s} (G_{2}'(F_{2}))^{T}
\end{gather}

while \textit{C}, $X_{1}$  come from source input $F_{1} $:
\begin{gather}\label{Eq 12}
X_{1} =G_{1}(F_{1}), C=f_{s} (G_{1}'(F_{1}))
\end{gather}

To provide a clear illustration of the interaction between $X_{1}$ and $X_{2}$, the above process can be expressed as matrix multiplication:
\begin{align}\label{Eq 13}
Y_{1} = C_{1} (B_{2}^{T} (X_{2}\cdot m_{2} )) + DX_{1} 
\end{align}

After that, the output $F_{1}^{c}$ of our SSD block is:
\begin{align}\label{Eq 14}
Z_{1} &=Linear(F_{1} ) \nonumber\\
F_{1}^{c} &=Z_{1} \times LN(Y_{1})
\end{align}

\begin{figure}[b] 
\centering
\includegraphics[width=0.45\textwidth]{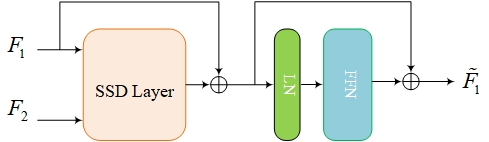}
\caption{Architecture of SSD blocks. A SSD block is composed of a SSD and a MLP layer. }
\label{Fig.3}
\end{figure}

SSD layer is shown as Fig. \ref{Fig.2}. Essentially, in Eq.(\ref{Eq 10}), our SSD first extract local enhanced structural feature of target by non-casual bi-directional scanning, then fuse source feature to target by matrix multiplication. $DX_{1}$ is used to supplement the local features of $F_{1}$. Compared with existing cross image interaction methods based on Mamba\cite{ref3},\cite{ref45}, SSD breaks the limitation of scale and protects the global context, which is additionally more computationally and training efficient than SSMs. Our SSD blocks follow the design of vanilla Transformers, which contains SSD token-mixer and feed forward layers. The architecture of SSD block is shown in Fig. \ref{Fig.3}. 

\begin{figure*}[t!]
\centering
\includegraphics[width=\textwidth]{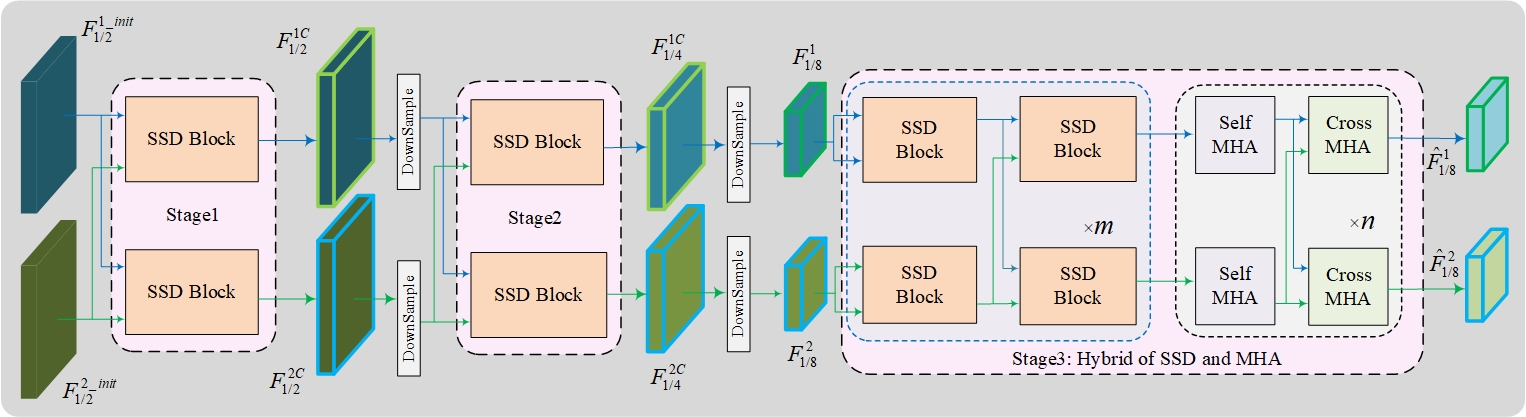}
\caption{Architecture of Cross-Modality feature Interaction (CMI) module. Stage1 and Stage2 are SSD blocks in cross form for feature extraction in scale of 1/2 and 1/4. Stage3 is hybrid of SSD and multi-head attention blocks applying in 1/8 scale. }
\label{Fig.4}
\end{figure*}

\subsubsection{CMI and MSF modules} 

CMI module is designed for shared structural feature extraction of two inputs in different scales, which is shown in Fig. \ref{Fig.4}. In LoFTR, XoFTR and other detector-free methods, the initial 1/2 and 1/4 features are extracted by CNN backbones. Since CNNs lacks global representation and cross-modal interaction, we introduce SSD in scale of 1/2 and 1/4 instead of CNN backbones.

In Stage1, $F_{1/2}^{1C}$ and $F_{1/2}^{2C}$ are calculated as:
\begin{align}\label{Eq 15}
F_{1/2}^{1C} & = SSD(F_{1/2}^{1_{-}init}, F_{1/2}^{2_{-}init})\nonumber \\F_{1/2}^{2C} & = SSD(F_{1/2}^{2_{-}init}, F_{1/2}^{1_{-}init})
\end{align}

Same as Stage1, we can get $F_{1/4}^{1C}$ and $F_{1/4}^{2C}$ after Stage2:
\begin{align} \label{Eq 16}
&F_{1/4}^{1_{-}init} =Downsample(F_{1/2}^{1C})\nonumber\\
&F_{1/4}^{2_{-}init} =Downsample(F_{1/2}^{2C})\nonumber\\
&F_{1/4}^{1C} = SSD(F_{1/4}^{1_{-}init}, F_{1/4}^{2_{-}init})\nonumber\\
&F_{1/4}^{2C} = SSD(F_{1/4}^{2_{-}init}, F_{1/4}^{1_{-}init})
\end{align}

In Stage3, we apply hybrid of SSD and multi-head attention (MHA) in scale of 1/8. This hybrid approach allows us to enhance the network's adaptability across different models and further mitigate the over-smoothing risk in SSD. The ratio of SSD blocks and MHA is 4:1. Experiments demonstrate this combination achieves an efficient balance between speed and performance in all multi-modal image registration. Similar to self and cross multi-head attention, a group of SSD layers ($SSD_{G} (F1,F2)$) for self and cross interaction can be calculated as:
\begin{align} \label{Eq 17}
&F_{1m} = SSD(F_{1} ,F_{1} ),F_{2m} = SSD(F_{2} ,F_{2} ),\nonumber\\
&F_{1} = SSD(F_{1m} ,F_{2m} ),F_{2} = SSD(F_{2m} ,F_{1m} )
\end{align}

The process of a vanilla multi-head attention layer group ($MHA_{G} (F1,F2)$) can also follow Eq. (\ref{Eq 17}), which replaces SSD with MHA. First, we get initial 1/8 scale features $F_{1/8}^{1}$ and $F_{1/8}^{2}$ from downsampled $F_{1/4}^{1C}$ and $F_{1/4}^{2C}$ , then Stage3 transfers them into coarse features $\hat{F}_{1/8}^{1}$ and $\hat{F}_{1/8}^{2}$. This process can be calculated as follows, where m=4, n=1:

\begin{align} \label{Eq 18}
&F_{1/8}^{1m} ,F_{1/8}^{2m} = (SSD_{G}(F_{1/8}^{1},F_{1/8}^{2})_{m},\nonumber\\
&\hat{F}_{1/8}^{1}, \hat{F}_{1/8}^{2} = (MHA_{G}(F_{1/8}^{1m},F_{1/8}^{2m}))_{n}
\end{align}

\begin{figure}[t!] 
\centering
\includegraphics[width=0.48\textwidth]{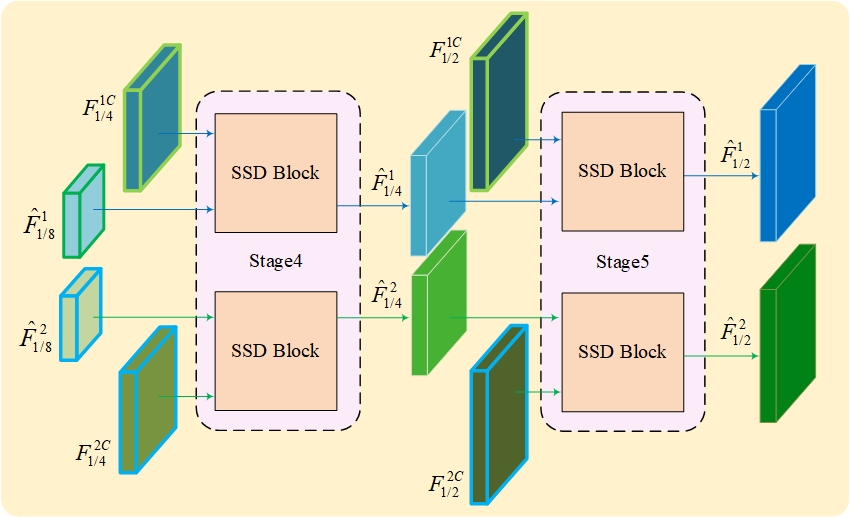}
\caption{Architecture of Multi-Scale feature Fusion (MSF) module. Stage4 and Stage5 are designed to fuse features in scale of [1/8,1/4] and [1/4,1/2] based on SSD.}
\label{Fig.5}
\end{figure}

After Stage3, we gain the coarse features in 1/8 scale for coarse matching. Following CMI module, MSF module is designed to fuse features in the scale of 1/8, 1/4, 1/2 progressively by SSD, in order to improve the robustness of noise and scale difference in multi-modal image registration. In traditional multi-scale feature fusion methods, feature map in smaller scale will be first upsampled by interpolation or transposed convolution, then fused into larger scale by addition or concatenation. However, this process will unavoidably introduce artifacts to edge and structural features, severely degrading the feature extraction in image registration. Therefore, we adopt SSD to feature fusion process and propose our novel MSF module. As mentioned in Section \ref{sec:block}, SSD first extracts structural features in smaller scale by bidirectional scanning and then fuses them to the features in a larger scale by matrix multiplication, without introducing irrelevant artifacts. 

Our MSF module is shown in Fig. \ref{Fig.5}. In Stage4, the fusion process of features in scale of 1/8 and 1/4 can be calculated as follows:
\begin{align} \label{Eq 19}
&F_{1/4}^{1} =SSD(F_{1/4}^{1C},\hat{F}_{1/8}^{1} )\nonumber\\
&F_{1/4}^{2} =SSD(F_{1/4}^{2C},\hat{F}_{1/8}^{2} )
\end{align} 

Same as Stage4, Stage5 fuses the features in the scale of 1/4 and 1/2, as following process:
\begin{align}\label{Eq 20}
&F_{1/2}^{1} =SSD(F_{1/2}^{1C},\hat{F}_{1}^{1/4} )\nonumber\\
&F_{1/2}^{2} =SSD(F_{1/2}^{2C},\hat{F}_{2}^{1/4} )
\end{align} 

After Stage5, we get the fine features in 1/2 scale, which is prepared for the last refinement and fine matching.

\subsection{Coarse-to-fine Matching}
Following \cite{ref30}, in coarse matching block, we calculate similarity matrix $\textit{S}_{c}$ of $\hat{F}_{1/8}^{1}$ and $\hat{F}_{1/8}^{2}$. After that, we multiplies the Softmax results of column and row to get matching probability. By argmax operation on probability scores, we obtain coarse correspondence $\textit{idx}_{1}$ and $\textit{idx}_{2}$.

In Fine Feature Matching Block, we gather 5×5 windows from fine feature maps $\hat{F}_{1/2}^{1}$ and $\hat{F}_{1/2}^{2}$ according to each coarse matching index and employ window MHA blocks. After that, we get the last refined features $f_{1}^{5\times 5}$ and $f_{2}^{5\times 5}$ of each patch. Like coarse matching, we calculate similarity matrix $S_{f}$ and multiplies the Softmax results of two dimensions. By argmax operation, we obtain fine correspondence $corr_{1}$ and $corr_{2}$.

Similar to other semi-dense matching methods, our loss function also contains coarse matching loss and fine matching loss, as follows:
\begin{align}\label{Eq 21}
L_{c} & = CE(\frac{S_{c} }{\tau } ,GT_{c} )\\
L_{f} & = \frac{1}{N} \sum_{i=1}^{N} CE(\frac{S_{f}^{i} }{\tau } ,GT_{f}^{i} ) 
\end{align}

CE is cross-entropy loss, $S_{c}$ is coarse similarity matrix, $S_{f}^{i}$ is fine similarity matrix of the i-th window, $GT_{c}$ is the ground truth of coarse correspondences while $GT_{f}^{i}$ is fine matching ground truth. Total loss is calculated as follow:
\begin{gather} \label{Eq 22}
L = L_{c}  + \alpha L_{f} 
\end{gather}

In our experience, we set $\alpha =2,\tau = 0.1$ .

\section{experience and results}
\subsection{Datasets and Evaluation Protocol}
\subsubsection{Datasets}
We have verified our method on the following three multi-modal image datasets: VIS-SAR, VIS-IR and VIS-NIR.

\textbf{VIS-SAR}. OSDataset \cite{ref39} is a VIS-SAR dataset and provides aligned SAR and optical image pairs with the size of 512×512 and 1m resolution, which are collected from GaoFen3 and Google Earth. OSDataset has 2011 pairs in training dataset and 424 pairs in test dataset.

\textbf{VIS-IR}. Same as ReDFeat \cite{ref19} and XoFTR \cite{ref30}, we use the following two aligned IR datasets: RGB-LWIR dataset in LGHD \cite{ref40} with 44 pairs and RoadSence in FusionDN \cite{ref41} with 221 pairs. The test set consists of all image pairs from LGHD and 47 randomly selected pairs from RoadSence, while the last 174 pairs for training.

\textbf{VIS-NIR}. RGB-NIR dataset \cite{ref42} has 477 RGB and NIR images with the wavelength of 750-1100nm. All pairs are registered which covering 9 different senses. We split the whole dataset to 345 pairs for training and 132 pairs for testing. 
\subsubsection{Evaluation metrics}
For qualitative evaluation of image registration performance, the following metrics are employed.

\textit{RMSE}: Root mean square error (RMSE) can be calculated as following:
\begin{align} \label{Eq 23}
RMSE = \sqrt{\frac{1}{N}\sum_{i=1}^{N}(H_{e}M^{i}(x,y)-H_{gt}M^{i}(x,y))}
\end{align}
where $H_{e}$is the homography transform which is estimated by MAGSAC \cite{ref43} method with reprojection threshold of 2 and the iterations of $1\times 10^{6} $, $H_{gt}$ is the ground-truth homography transform, \textit{M} is points set, which is uniformly selected on the image with the step of $(H/32, W/32)$, N=961.

\textit{NCM}: Number of corrected matching points (NCM) is defined as the number of all correspondences with the distance less than 2 pixels to ground truth. 

\textit{SMR}: Successful matching rate (SMR) is the proportion of the number of matching pairs with the RMSE lower than 5 relative to the complete test set. 
\subsection{Experimental setup}

Our experiments are operated based on Pytorch platform. In training strategy, optimizer is AdamW \cite{ref44} with a cosine annealing scheduler. The initial learning rate (LR) is $1\times  10^{-3} $ for pretraining and $1\times  10^{-4} $ for finetuning and the minimum of LR is set to $1\times  10^{-6} $. For our RegNetMamba-2 and other semi-dense methods in this paper, we first pretrain them on COCO and finetune on three multi-modal training datasets. Before training, images are randomly transformed with rotation of [$-40^{\circ}$ , $40^{\circ}$] and scale of [0.8,1.2]. Our batch size is set to 1 both in pretraining and finetuning.

The dimensions of our network are [128, 192, 256, 192, 128], while the size of feature maps in Stage1 to Stage5 are [1/2, 1/4, 1/8, 1/4, 1/2], and the head of MHA in coarse and fine level is [8,4].
\begin{table}[b]
\caption{Ablation study of SSD and local enhancement for image registration on VIS-SAR, VIS-IR and VIS-NIR.\label{tab.1}}
\centering
\begin{minipage}{0.5\textwidth}
\centering
\begin{tabular}{|c|c|c|c|c|}
\hline
    Dataset & Method & aRMSE↓ &	aNCM↑ & SMR↑ \\
\hline
\multirow{3}{*}{VIS-SAR} & LA-wo LE	& 2.50 & 218.48 & 0.901 \\
                &  SSD-wo LE & 2.33 & 442.16 & 0.920 \\
                &  SSD + LE & \textbf{2.28} & \textbf{487.07} & \textbf{0.934} \\
\hline
\multirow{3}{*}{VIS-IR} &LA-wo LE & 2.35 & 371.24 & 0.977\\
&SSD-wo LE & 2.10 & 437.36 & 0.977\\
&SSD + LE & \textbf{1.97} & \textbf{589.80} & \textbf{0.977}\\
\hline

\multirow{3}{*}{VIS-NIR}& LA-wo LE & 0.82 & 3542.98 & 0.955\\
& SSD-wo LE & 0.79 & 3915.25 & 0.955\\
& SSD + LE & \textbf{0.72} & \textbf{3992.64} & \textbf{0.955}\\
\hline

\end{tabular}
\end{minipage}
\end{table}
\subsection{Ablation Study}
\subsubsection{Ablation study of SSD and local enhancement}
Current semi-dense matching methods are mostly based on vanilla attention or linear attention in coarse and fine level. To verify whether SSD performs better than linear attention, in experience (1) we replace all SSD blocks with linear attention (LA) in our architecture for baseline. Another modification is feature scaling on B, C for local representation enhancement (called LE). To test the effectiveness, experience (2) SSD-wo LE is vanilla SSD without local enhanced; experience (3) is the whole method SSD + LE. 

The results of experiences are shown in Tabel. \ref{tab.1}. Under the same architecture, SSD performs better than linear attention in VIS-SAR with a higher aNCM, SMR and lower aRMSE. Although speckle noise in SAR degrades the structural and edge information of images, SAR offers the advantages of high resolution and high contrast. Vanilla SSD shows excellent global structural 
feature extraction with superior denoising capability. In VIS-IR image dataset, some optical images are captured under night-time or overexposed conditions, making foreground features difficult to separate from the background. Compared to linear attention, vanilla SSD performs also better than linear attention, since SSD is inherently better than linear attention at extracting the main foreground information. For VIS-NIR, results of two methods have marginal disparity due to the minimal noise and distinct textures. 

To fully leverage the global representation of SSD, while enhancing its foreground edge and structural feature extraction performance, we introduce feature scaling to enhance local representation of SSD. Our local enhanced SSD in VIS-SAR, VIS-IR and VIS-NIR image registration achieves the best results across all ablation metrics, which has the smallest aRMSE, the highest aNCM and SMR. Compared with vanilla SSD, in VIS-SAR, our method gets 0.05↓ of aRMSE, 44.91↑ of aNCM and $1.4\%$↑  of SMR; in VIS-IR, we get 0.13↓ and 152.44↑ of aRMSE and aNCM; in VIS-NIR, we gain 0.07↓ of aRMSE and 77.39↑ of aNCM. This result verifies that feature scaling function of local enhancement has improved the registration performance of SSD with only a modest increase in computational cost, effectively enhancing its capability to extract local features, thereby enabling adaptation to all different modalities.
\begin{table}[t!]
\caption{Ablation study of CMI and MSF module for image registration on VIS-SAR, VIS-IR and VIS-NIR.\label{tab.2}}
\centering
\begin{minipage}{0.5\textwidth}
\centering
\begin{tabular}{|c|c|c|c|c|}
\hline
    Dataset & Method & aRMSE↓ &	aNCM↑ & SMR↑ \\
\hline
\multirow{3}{*}{VIS-SAR} & baseline	& 2.49 & 366.42 & 0.902 \\
                &  only CMI & 2.32 & 477.12 & 0.920 \\
                &  only MSF & 2.38 & 435.38 & 0.908 \\
                & CMI + MSF & \textbf{2.28} & \textbf{487.07} & \textbf{0.934} \\
\hline
\multirow{3}{*}{VIS-IR} & baseline	& 2.19 & 350.45 & 0.977 \\
                &  only CMI & 2.03 & 444.18 & 0.977 \\
                &  only MSF & 2.16 & 389.80 & 0.977 \\
                & CMI + MSF & \textbf{1.97} & \textbf{589.80} & \textbf{0.977} \\
\hline
\multirow{3}{*}{VIS-NIR} & baseline	& 0.81 & 3781.80 & 0.955 \\
                &  only CMI & 0.76 & 3915.25 & 0.955 \\
                &  only MSF & 0.80 & 3878.21 & 0.955 \\
                & CMI + MSF & \textbf{0.72} & \textbf{3992.64} & \textbf{0.955} \\
\hline

\end{tabular}
\end{minipage}
\end{table}
\subsubsection{Ablation study of CMI and MSF module}
In order to verify the contribution of CMI and MSF modules, we set the following four experiences: 

(1). Similar to the architecture of LoFTR, we only maintain the Stage 3 in the scale of 1/8, while the features in 1/4 and 1/2 are extracted by CNN without multi-scale feature fusion. We set this experience as baseline.

(2). Based on experience (1), we only apply CMI module for feature extraction in scale of 1/2, 1/4 and 1/8 without fusion. 

(3). The multi-scale features are extracted by CNNs. We only retain MSF module for feature fusion in Stage4 and Stage5.

(4). The whole architerture of RegNetMamba-2 with CMI and MSF modules.

Table. \ref{tab.2} shows the results of the four ablation experiences. Compared with baseline, CMI module plays more important part than MSF. In VIS-SAR, CMI modules gets 0.17↓ of aRMSE and 110.70↑ of aNCM, while MSF modules gets 0.11↓ and 68.96↑ of aRMSE and aNCM. In VIS-IR, CMI gets 0.16↓ and 93.78↑ of aRMSE and aNCM, while MSF gets 0.03↓ and 39.35↑. In VIS-NIR, CMI gains 0.05↓ of aRMSE and 133.45↑ of aNCM, while the improvement of MSF in aRMSE and aNCM is 0.01↓ and 96.41↑. This result verifies that CMI modules provides better shared structural features in scale of 1/2, 1/4 and 1/8 than vanilla CNNs, and the following MSF module further aggregates features from different receptive fields. RegNetMamba-2 gets the best performance in the registration of VIS-SAR, VIS-IR and VIS-NIR image registration benefitting from combination of CMI and MSF modules.

\begin{table}[b!]
\caption{Average RMSE, NCM and SMR of RegNetMamba-2 and 7 comparison methods for image registration on three datasets: VIS-SAR, VIS-IR and VIS-NIR.}
\label{tab.3}
\centering
\begin{minipage}{0.5\textwidth}
\centering
\begin{tabular}{|c|c|c|c|c|}
\hline
    Dataset & Method & aRMSE↓ &	aNCM↑ & SMR↑ \\
\hline
\multirow{6}{*}{VIS-SAR} & RIFT	& 3.67 & 59.18 & 0.104 \\
                &  Cnet & 2.94 & 110.25 & 0.630 \\
                & ReDFeat & 2.59 & 273.27 & 0.795 \\
                & LoFTR & 2.39 & 112.05 & 0.820 \\
                & XoFTR & \underline{2.33} & 233.06 & 0.863 \\
                & LoFLAT & 2.69 & 304.40 & 0.884 \\
                & JamMa & 2.55 & \underline{396.23} & \underline{0.906} \\
                & \textbf{RegNetMamba-2} & \textbf{2.28} & \textbf{487.07} & \textbf{0.934}\\
\hline
\multirow{6}{*}{VIS-IR} & RIFT	& 3.71 & 245.71 & 0.361 \\
                &  Cnet & 3.01 & 112.11 & 0.721 \\
                & ReDFeat & \underline{2.44} & 203.32 & 0.907 \\
                & LoFTR & 2.78 & 388.25 & 0.872 \\
                & XoFTR & 2.53 & \underline{555.60} & 0.883 \\
                & LoFLAT & 2.67 & 270.15 & 0.930 \\
                & JamMa & 2.65 & 450.37 & \underline{0.977}\\
                & \textbf{RegNetMamba-2} & \textbf{1.97} & \textbf{589.80} & \textbf{0.977}\\
\hline

\multirow{6}{*}{VIS-NIR} & RIFT	& 1.31 & 1076.63 & 0.886 \\
                &  Cnet &   1.02 &  1204.24 & 0.947\\
                & ReDFeat & 0.86 & 2707.43 & 0.947 \\
                & LoFTR & 0.79 & 3523.24 & 0.955 \\
                & XoFTR & \underline{0.74} & \underline{3791.52} & \underline{0.955} \\
                & LoFLAT & 0.77 & 3604.90 & 0.955 \\
                & JamMa & 0.75 & 3643.58 & 0.955\\
                & \textbf{RegNetMamba-2} & \textbf{0.72} & \textbf{3992.64}& \textbf{0.955}\\
\hline

\end{tabular}
\end{minipage}
\end{table}
\subsection{Image Registration Performance}
\subsubsection{Comparison Methods setup}
To validate the effectiveness of our approach, we conduct comparative experiments with several representative methods across different categories. As detector-based method, we select RIFT \cite{ref9}, Cnet \cite{ref12} and ReDFeat \cite{ref19} as comparation, which represent handcrafted method, patch-based deep learning method and joint detection and description methods. For RIFT, we select 5000 keypoints by FAST detector on maximum momentum and extract descriptors on the 96×96 patches of MIM feature map. For Cnet, we adopt the same keypoints detection as RIFT to select 5000 keypoints, and image patch size is 64×64. For ReDFeat, we apply non-maximum suppression with the kernel size of 7×7 on score maps for feature detection and select top 5000 keypoints with the highest scores. For detector-free methods, we choose LoFTR \cite{ref26} ,XoFTR \cite{ref30}, LoFLAT\cite{ref47} and JamMa\cite{ref45} for comparison. Due to varying image characteristics, it is difficult to determine a fixed matching threshold for all datasets. Instead, we select correspondences with the top 5000 matching scores in fine matching level for all semi-dense methods to ensure a consistent number of feature points across comparative methods.
\subsubsection{Performance of image registration on three datasets}

\begin{figure}[b!] 
\centering
\includegraphics[width=0.45\textwidth]{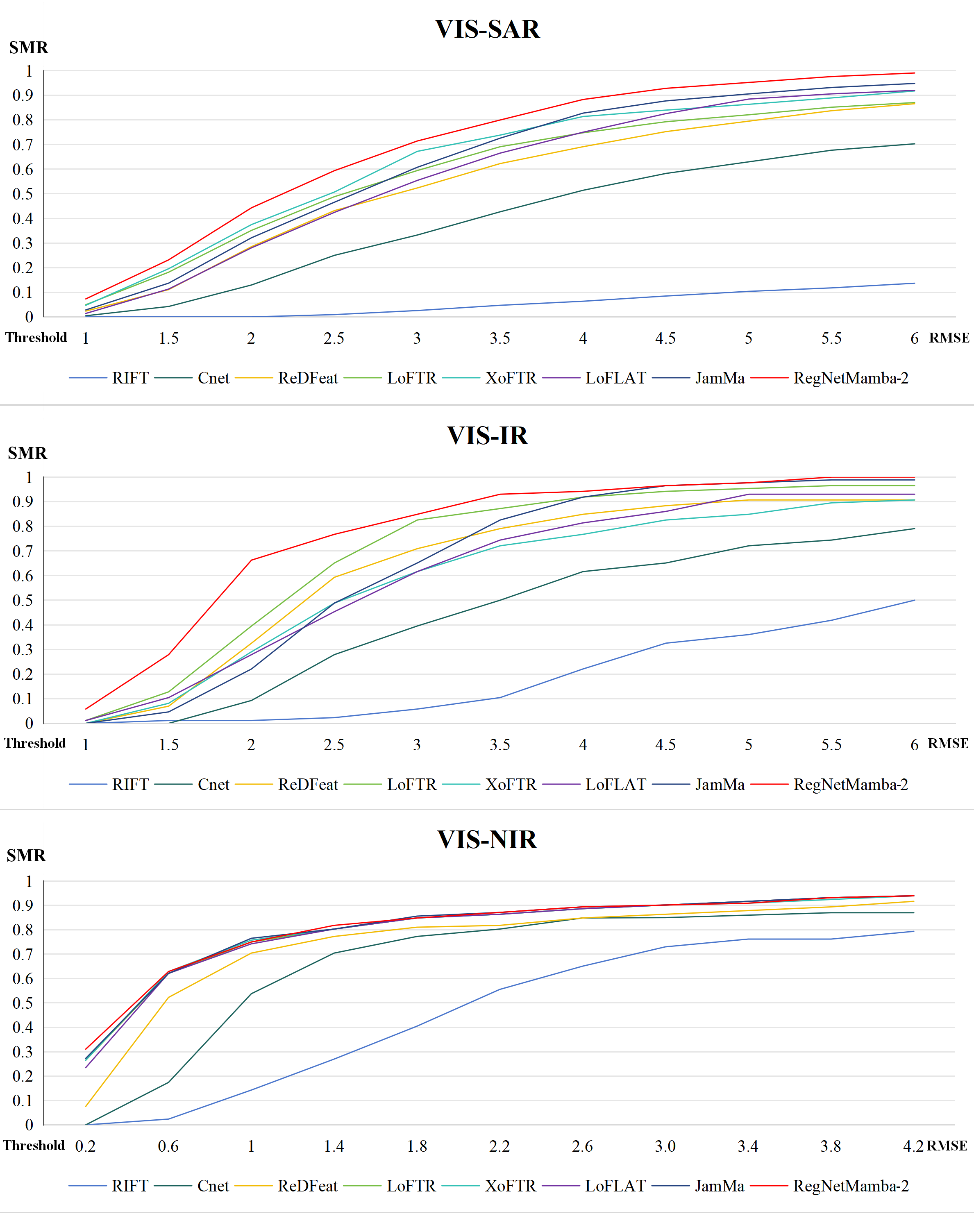}
\caption{Successful matching rate (SMR) of RIFT, Cnet, ReDFeat, LoFTR, XoFTR, LoFLAT, JamMa and our RegNetMamba-2 at varying thresholds of RMSE up to 6 for VIS-SAR, VIS-IR and to 4.2 for VIS-NIR. The X-axis is the threshold, while the Y-axis displays the SMR of image registration.}
\label{Fig.6}
\end{figure}

\begin{table*}[t!]
\caption{RMSE and NCM of RegNetMamba-2 and 5 comparison methods for image registration on six image pairs.\label{tab.4}}
\centering
\begin{minipage}{\textwidth}
\centering
\begin{tabular}{|c|c|c|c|c|c|c|c|c|c|c|c|c|}
\hline
 Pairs & \multicolumn{2}{c|}{Pair1} & \multicolumn{2}{c|}{Pair2}& \multicolumn{2}{c|}{Pair3} & \multicolumn{2}{c|}{Pair4}& \multicolumn{2}{c|}{Pair5} & \multicolumn{2}{c|}{Pair6}\\
\hline
     Method &	NCM↑ & RMSE↓ &	NCM↑ & RMSE↓ &	NCM↑ & RMSE↓ &	NCM↑ & RMSE↓ &	NCM↑ & RMSE↓ &	NCM↑ & RMSE↓ \\
\hline
RIFT	& 32 & 10 & 9 & 10 & 23 & 10 & 10 & 10 & 136 & 7.65 & 57 & 8.02\\
Cnet & 12 & 10 & 16 &10 & 13 & 10 & 8 & 6.50& 376 & 1.13 & 177 &4.24 \\
ReDFeat  & 50 & 10 & 25 & 7.23 & 25 & 5.67 & 55 & \underline{2.30} & 1539 & 0.35 & 960 & 1.18\\
LoFTR  & 35 & 10 & 68 & \underline{1.88} & 59 & 10 & 114 & 2.31& 1044 & 0.38 & 3264 & \underline{1.05} \\
XoFTR & 117 & \underline{1.92} & 102 & 2.09 & 127 & 6.01 & 166 & 2.55& 2232 & 0.40 & \underline{3873} & 1.15 \\
LoFLAT & 116 & 2.00 & 92 & 4.88 & \underline{132} & 6.19 & 212 & 2.36 &  1526 & 0.32 & 2521 & 1.44 \\
JamMa & \underline{245} & 3.40 & \underline{203} & 3.38 & 122 & \underline{4.38} & \underline{224} & 2.65 & \underline{2501} & \underline{0.26} & 3203 & 1.22 \\
RegNetMamba-2 & \textbf{254} & \textbf{1.80} & \textbf{378} & \textbf{1.63} & \textbf{330} & \textbf{1.79} & \textbf{228} & \textbf{1.76} & \textbf{3844} & \textbf{0.19} & \textbf{4119} & \textbf{0.78} \\
\hline
\end{tabular}
\end{minipage}
\end{table*}

\begin{figure*}[b!]
\centering
\includegraphics[width=0.95\textwidth]{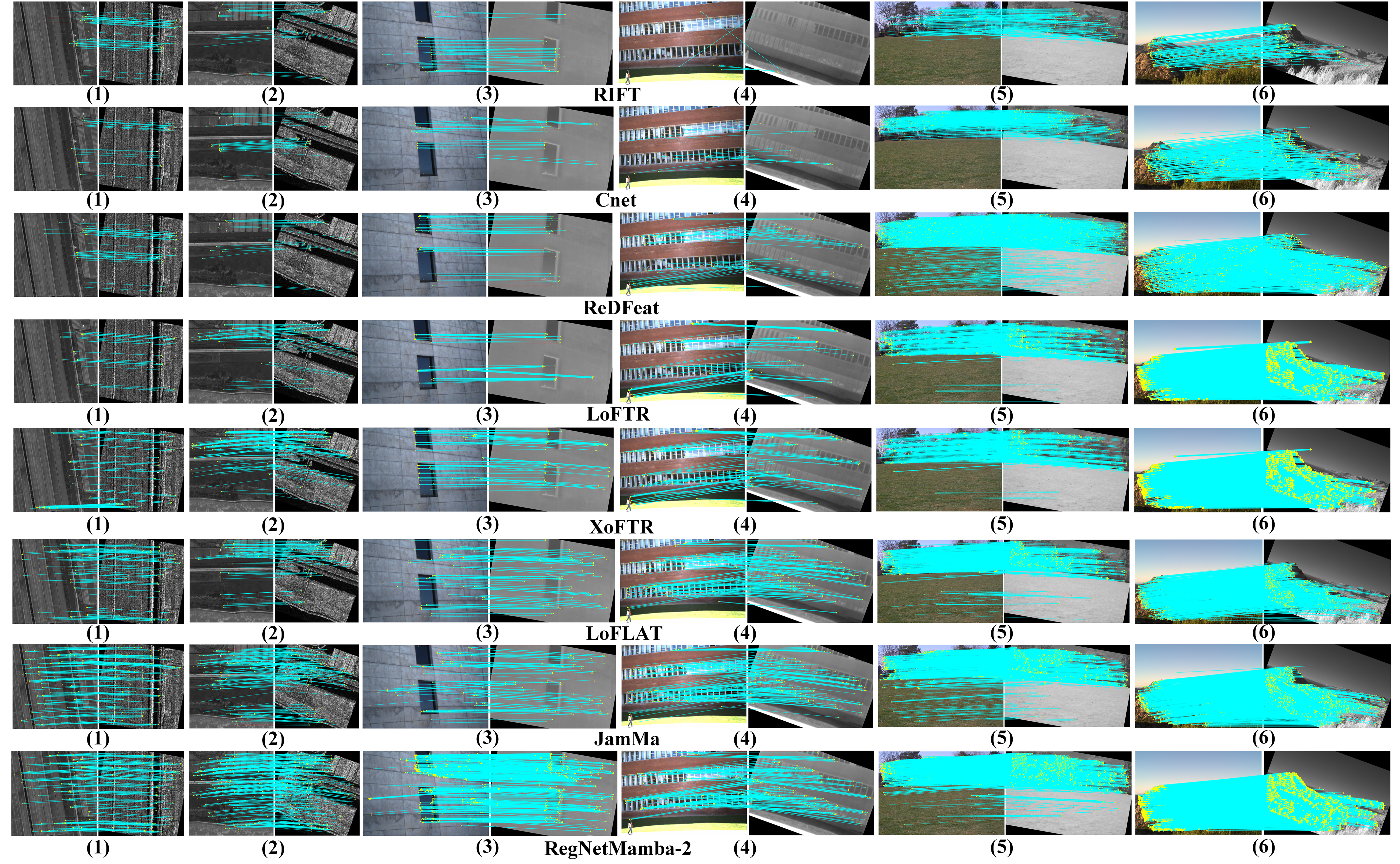}
\caption{Visualization of matching performance. Only the inlier matches after MAGSAC are shown.}
\label{fig.7}
\end{figure*}

\begin{figure*}[b!]
\centering
\includegraphics[width=0.95\textwidth]{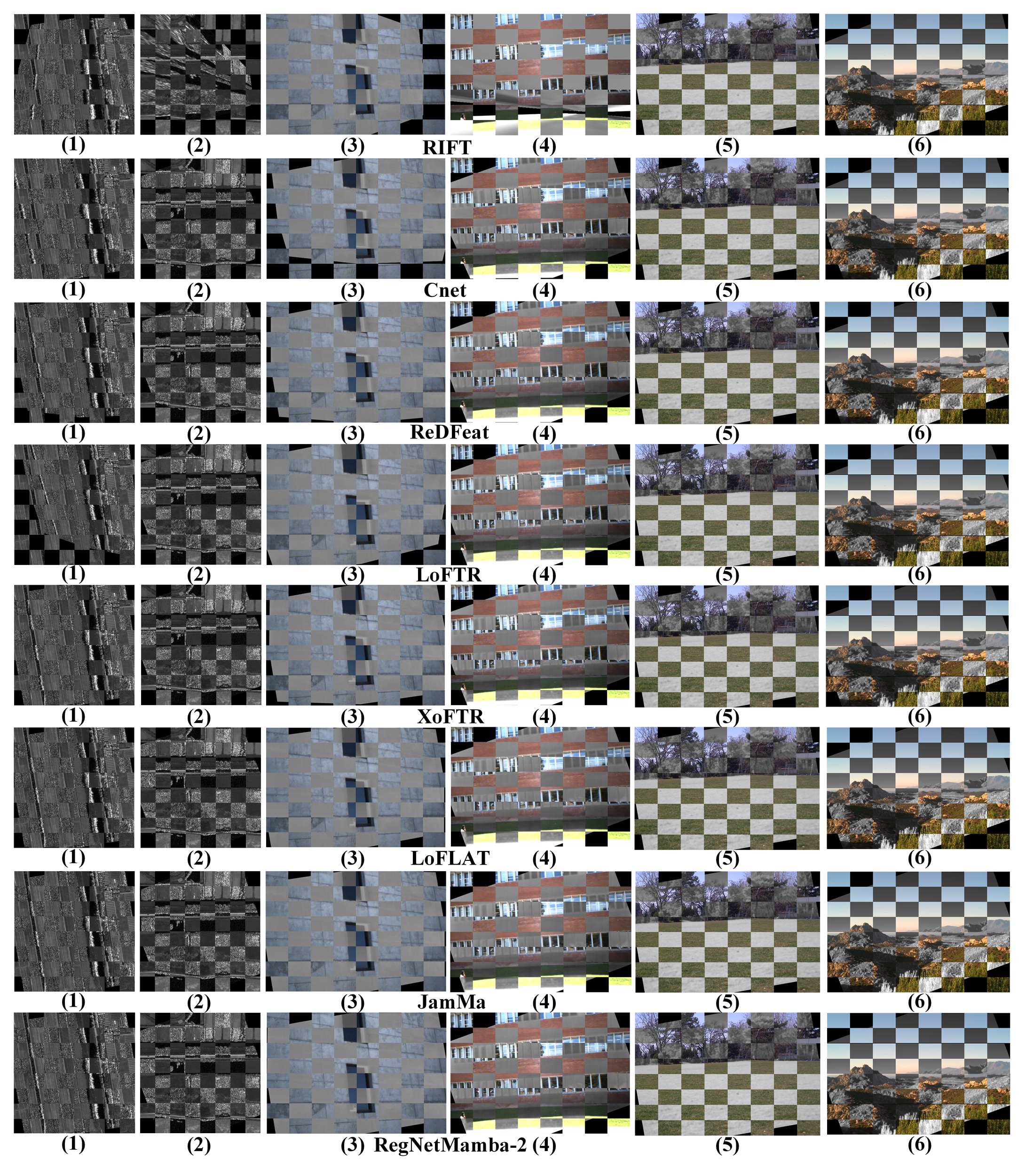}
\caption{Checkerboard images of Pair1 to Pair 6 registration results.}
\label{fig.8}
\end{figure*}

\begin{figure*}[b!]
\centering
\includegraphics[width=0.95\textwidth]{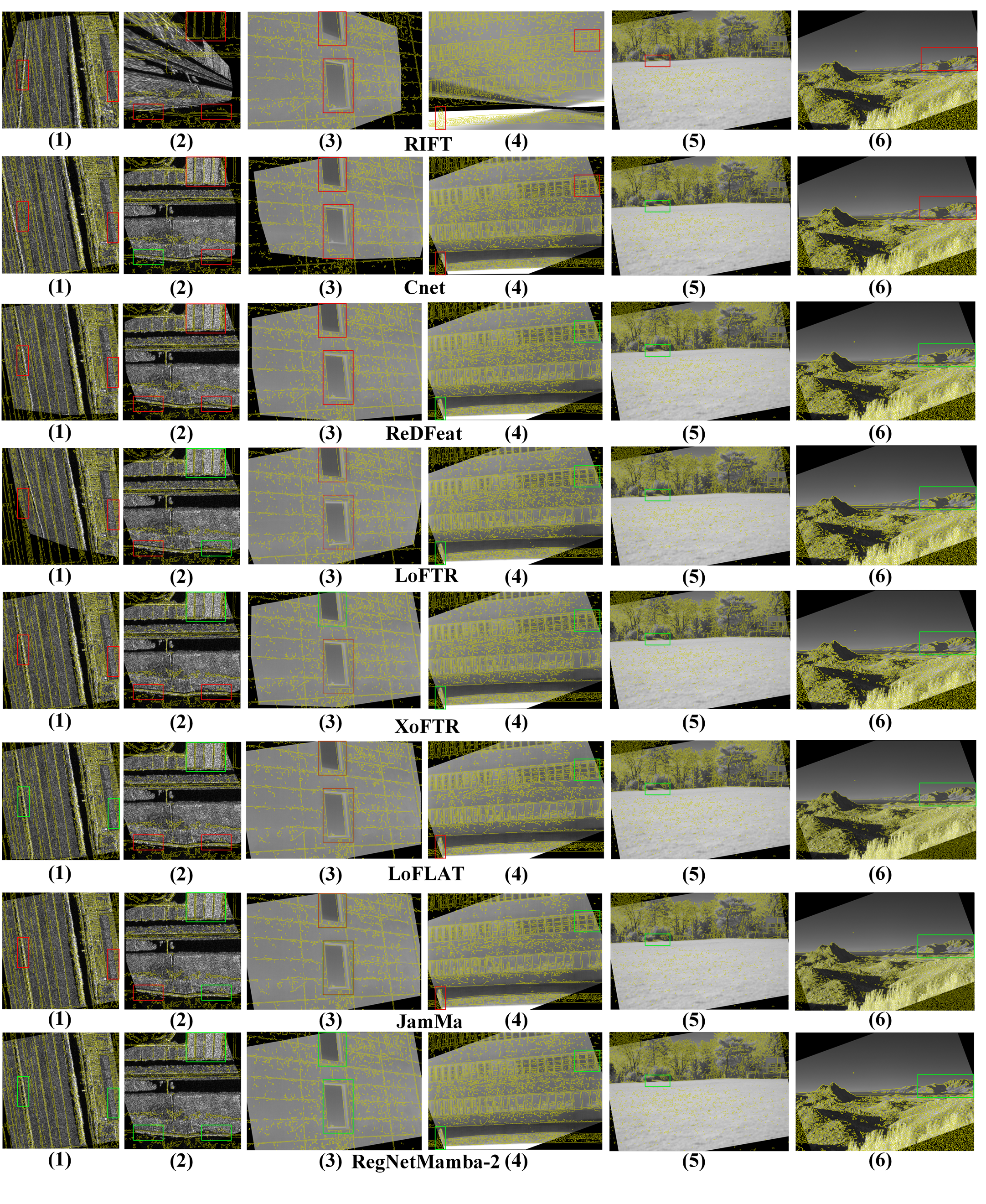}
\caption{Edge images of Pair1 to Pair 6 registration results. Red and green rectangles circle the areas not aligned or aligned.}
\label{fig.9}
\end{figure*}

Table. \ref{tab.3} presents the performance comparison of RIFT, Cnet, ReDFeat, LoFTR, XoFTR, LoFLAT, JamMa and RegNetMamba-2 on the VIS-SAR, VIS-IR, and VIS-NIR datasets. For detector-based methods, ReDFeat delivers the second best performance across the VIS-SAR and VIS-IR datasets in aRMSE. However, as a CNN-based method, ReDFeat remains limited in global representation and noise suppression. For detector-free methods, LoFTR and XoFTR employ self and cross attention in coarse-level, enabling a global receptive field and cross-modal interaction. XoFTR further enhances fine-level matching by incorporating multi-scale feature fusion building upon LoFTR’s architecture, which helps it outperform other comparison methods. LoFLAT introduced focus function to linear attention, but linear attention performs not as well as SSD especially for multi-modal images. JamMa first introduces SSM into image matching, but the joint scan path of two images inevitably destroys the global structural features of an entire image and introduces irrelevant context. Oppositely, our SSD in cross form essentially displays a process of obtaining the global features from another image by scanning, then delivering the whole global information to the source image, thus effectively preserving the continuity of features in interaction of two inputs. Furthermore, RegNetMamba-2 designs CMI and MSF module, significantly improving local and structural representation as well as robustness of noise in different scales.

RegNetMamba-2 is a novel detector-free semi-dense matching algorithm, achieving the best performance across all three datasets. Compared with the second best results in VIS-SAR dataset, aRMSE of RegNetMamba-2 is 0.05↓ than XoFTR, while aNCM and SMR is 90.84↑ and 2.8\%↑ compared with JamMa. For VIS-IR dataset, we can get 0.47↓ of aRMSE than ReDFeat, with 34.2↑ of aNCM than JamMa. For VIS-NIR dataset, we have obtained 0.02↓ and 201.12↑ of aRMSE and aNCM than XoFTR.

Fig. \ref{Fig.6}. illustrates the successful matching rate (SMR) for 5000 keypoints across all algorithms under varying thresholds of RMSE in VIS-SAR, VIS-IR, and VIS-NIR registration tasks. RegNetMamba-2 consistently achieves the highest SMR across all thresholds of multi-modal datasets. In VIS-SAR registration, our method outperforming all seven baseline methods and achieves to 100\% SMR at threshold 6. For VIS-IR, our advantages are more obvious. RegNetMamba-2 reaches 100\% SMR at threshold 5 and RMSE of most samples are lower than 3.5 in VIS-IR registration. For VIS-NIR, where images contain richer details and less noise, all semi-dense methods perform competitively, though our approach still slightly surpasses the others.

\subsubsection{Performance of image registration on six image pairs}
We select six pairs from three datasets, while Pair 1 and Pair 2 are from VIS-SAR, Pair3 and Pair 4 are from VIS-IR, Pair 5 and Pair 6 are from VIS-NIR. All image pairs have been transformed before image registration. Tabel \ref{tab.4} shows the performance of RegNetMamba-2 and other 7 comparison by displaying NCM and RMSE of 6 pairs.

As shown in Tabel \ref{tab.4}, when the method fails to match image pairs, we set the RMSE up to 10.  Fig. \ref{fig.7} displayed the matching performance of these methods. Fig. \ref{fig.8} and Fig. \ref{fig.9} shows the checkerboard and edge images for measuring the accuracy of image registration. Red rectangles indicate the areas that are not aligned, and the green rectangles indicate the aligned areas.

For VIS-SAR image pairs, although both image pairs contain prominent structural information, the SAR images are affected by severe speckle noise. PC-based detector in RIFT and Cnet fails to extract enough accurate keypoints, while noise in patches reduces the discriminability of the descriptor, leading to numerous mismatches. ReDFeat also struggles to match both VIS-SAR pairs, since CNN backbone exhibits limited noise robustness and global representation. LoFTR achieves good results on Pair 2 but fails on Pair 1. The standard self and cross attention in LoFTR aggregate all tokens, inevitably introducing noise into descriptors especially during fine-level matching. XoFTR performs better in two pairs, since multi-scale fusion model partially suppresses noise and integrates global context into local matching. LoFLAT performs worse in Pair 2, due to the feature extraction capability of linear attention is limited. Even with the introduction of a focus function, it remains difficult to accurately focus on texture and structural features when facing images with poor details and heavy noise. For JamMa,the scanning direction of SSM reveals the weakness of the interaction capability and the disruption of global feature consistency, leading to poor results. In addition to the shortcomings of the backbone, the use of only single-scale feature extraction and interaction also results in insufficient capability of current mainstream semi-dense matching methods in extracting shared structural features. Our RegNetMamba-2 replaces standard attention with local enhanced SSD in CMI and MSF module, significantly enhancing local and structural feature extraction with superior robustness to speckle noise. 

For VIS-IR image pairs, Pair 3 represents a challenging case characterized by blurred edges and a critical lack of local detail information, which leads to persistent difficulties in accurately localizing keypoints and extracting distinguishable descriptors. All detector-based methods fail to achieve correct matching results under these conditions. Detector-free methods circumvent keypoint detection, but their attention or SSM still struggle to produce sharp distributions in low-texture regions. Focus function in flatten linear attention in LoFLAT is also hard to face with this problem. On the other hand, the encoder of LoFTR, XoFTR, LoFLAT, JamMa is CNNs, leading to the lack of multi-scale feature extraction and interaction. In contrast, our method benefits from the architecture design of CMI and MSF as well as SSD's inherent ability to extract forehead structural features. With local feature enhancement, SSD can more effectively extract highly discriminative features and leads to a higher number of accurate correspondences. Pair 4 presents a more conventional sample with a $20^{\circ}$ rotation. Our method attains lower RMSE and larger NCM values. RIFT continues to struggle with larger rotations, since MIM feature maps contain only six orientations. Cnet also yields higher pixel errors and lower aNCM, owing to the inherent limitations of vanilla CNNs in handling rotational variations.

\begin{table*}[t!]
\caption{Runtime and average RMSE of feature extraction in six methods. the runtime is counted in millisecond (ms).}
\label{tab.5}
\centering
\begin{minipage}{\textwidth}
\centering
\begin{tabular}{|c|c|c|c|c|c|c|c|}
\hline
\multirow{2}{*}{\diagbox{Methods}{Datasets}} & \multicolumn{2}{c|}{VIS-SAR} & \multicolumn{2}{c|}{VIS-IR} & \multicolumn{2}{c|}{VIS-NIR}\\
\cline{2-7}
      & aRMSE↓ & Time↓ & aRMSE↓ & Time↓  & aRMSE↓ & Time↓\\
\hline
RIFT & 3.67 & 5302  & 3.71 & 6107 & 1.31  &9840 \\
Cnet & 2.94 & 3969  & 3.01 & 4028 & 1.02 &5228  \\
ReDFeat & 2.59 & \textbf{197}   & \underline{2.44} &\textbf{200} & 0.86 &\textbf{295}\\
LoFTR & 2.39 & 227 & 2.78 & 238 & 0.79 &854 \\
XoFTR & \underline{2.33} & 251	& 2.53 & 269 & \underline{0.74} & 877 \\
LoFLAT & 2.69 & \underline{217} & 2.67 & \underline{226} & 0.77 & \underline{493} \\
JamMa & 2.55 & 263	& 2.65 & 276 & 0.75 & 569 \\
RegNetMamba-2 &\textbf{2.28} &339  & \textbf{1.97} & 349 & \textbf{0.72} & 696\\
\hline
\end{tabular}
\end{minipage}
\end{table*}

For VIS-NIR image pairs, the lower noise levels and richer texture and structural features allow most methods to achieve excellent matching performance. As semi-dense matching approaches, RegNetMamba-2 yield significantly more NCMs than detector-based methods, since they bypass the keypoint detection. In Pair 5, all methods successfully produce good registration results. As illustrated in Fig. \ref{fig.9}, the red rectangle highlights a region where RIFT shows noticeable misalignment, resulting in a higher RMSE. The other five methods all achieve good alignment. In Pair 6, where the NIR image is rotated by $20^{\circ}$, both RIFT and Cnet exhibit considerable performance degradation with higher RMSE, confirming their limited robustness to rotation which is consistent with the analysis mentioned in Pair 4.

\subsection{Efficiency Analysis}

To evaluate the performance and efficiency of feature extraction, we measure only the computation time required for feature detection and description in detect-then-describe methods, or the full network runtime in end-to-end methods, excluding time spent on feature matching. All experiments were conducted on an Intel i9-13900K CPU and an NVIDIA GeForce RTX 4090 24G GPU. RIFT runs entirely on the CPU in MATLAB, as it lacks GPU support. For Cnet, the handcrafted keypoint detector operates on the CPU, while descriptor extraction uses CNN on the GPU. All end-to-end methods as ReDFeat, LoFTR, XoFTR, and our RegNetMamba-2 are executed on the GPU. Table. \ref{tab.5} presents the runtime comparisons across the six methods. The input sizes of images are $512 \times 512$ for VIS-SAR, $640 \times 431$ for VIS-IR, and $1024 \times 715$ for VIS-NIR. 

RIFT and Cnet exhibit significantly higher computational costs. They both require to compute phase congruence and apply FAST detector operations which are not GPU-accelerated. Furthermore, Cnet processes 5000 patches in size of 64×64 by CNN, introducing substantial computational overhead. Even though RIFT and Cnet are time-consuming, they still struggle to achieve satisfactory registration results. 

ReDFeat achieves the fastest feature extraction speed by processing the entire image with a CNN rather than using a large number of patches. However, constrained by the limited receptive field of CNN and the lack of cross-modal interaction, ReDFeat demonstrates only moderate performance in SAR and IR image registration. Our method RegNetMamba-2 is slightly slower than other semi-dense methods at smaller image sizes, because they both adopt CNN for initial feature extraction, while SSD is applied in our whole architecture. However, our approach provides shared global structural features and achieves effective fusion of features in all scales, leading to superior performance in VIS-SAR and VIS-IR registration. For VIS-NIR images in larger size, our method performs better and faster than LoFTR and XoFTR, because they rely on softmax attention with O(N²) complexity, resulting in rapidly increasing computation at larger resolutions, while SSD operates with only linear O(N) complexity. Although LoFLAT and JamMa are also O(N) complexity and faster than us, RegNetMamba-2 performs better with a small additional time cost in CMI and MSF modules. In conclusion, RegNetMamba-2 achieves good effects in both performance and efficiency in registration of all multi-modal image datasets.

\section{conclusion}
In this paper, we incorporates SSD into coarse-to-fine semi-dense matching architecture for multi-modal image registration and propose our novel algorithm named RegNetMamba-2. Firstly, SSD is applied in different scales for multi-modal feature extraction through the whole network. To enhance local representation of SSD, we pay more attention on edge and structural features by feature scaling function of SSD. Secondly, we construct a novel cross-modality feature fusion model based on SSD, which contains CMI and MSF modules. CMI module is designed for shared feature extraction, which extends SSD into cross form and applies SSD in each scale. MSF module is designed for progressive upward fusion of multi-modal features in three different scales by SSD. Experiences in ablation studies demonstrate that SSD has a superior structural feature extraction capability with robustness against noise, while local enhancement significantly mitigates the over-smoothing of SSD. Comparison experiences in VIS-SAR, VIS-IR, VIS-NIR datasets and image pairs prove that RegNetMamba-2 achieves state-of-the-art results and exhibits exceptional cross-modal adaptability, while maintaining good effects in both performance and efficiency.

\begin{IEEEbiography}[{\includegraphics[width=1in,height=1.25in,clip,keepaspectratio]{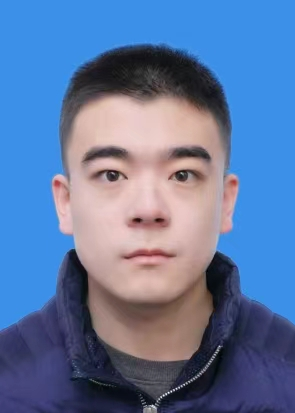}}]{Zhikang Li}received the B.S. degree from Xidian University, Xi’an, China, in 2021. He is currently pursuing the Ph.D. degree with the Remote Sensing Image Processing and Fusion Group, School of Electronic Engineering, Xidian University. 
	
His research interests include synthetic aperture radar image analysis and feature extraction.\end{IEEEbiography}

\begin{IEEEbiography}[{\includegraphics[width=1in,height=1.25in,clip,keepaspectratio]{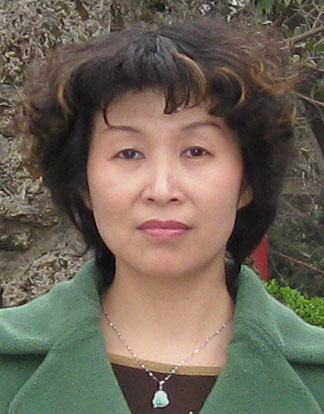}}]{Yan Wu}(Member, IEEE) received the B.S. degree	in information processing and the M.S. and Ph.D. degrees in signal and information processing from Xidian University, Xi’an, China, in 1987, 1998, and 2003, respectively.
	
From 2003 to 2005, she was a Post-Doctoral Fellow with the National Key Laboratory of Radar Signal Processing, Xidian University, where she has been a Professor with the Department of Electronic Engineering since 2005. She has published more than 80 technical articles. Her broad research inter-ests are remote sensing image analysis and interpretation, data fusion of multi-sensor images, synthetic aperture radar (SAR) autotarget recognition, and statistical learning theory and application.
\end{IEEEbiography}

\begin{IEEEbiography}[{\includegraphics[width=1in,height=1.25in,clip,keepaspectratio]{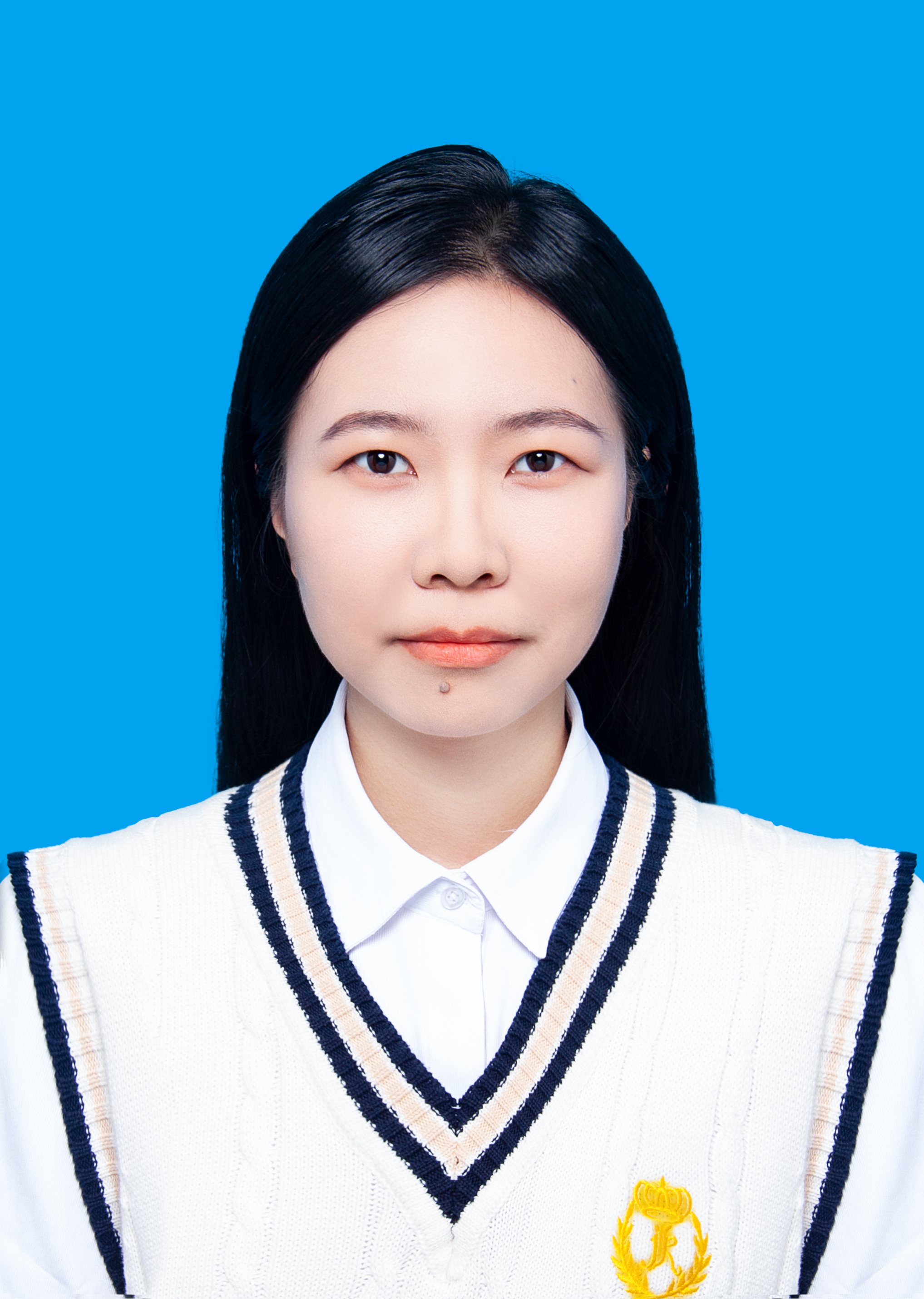}}]{Xin Hu} received the B.S. degree from the School of Automation and Information Engineering, Xi’an University of Technology, Xi’an, China, in 2019. She is currently pursuing the Ph.D. degree with the Remote Sensing Image Processing and Fusion Group, School of Electronic Engineering, Xidian University, Xi’an.
	
Her main research direction is multimodal remote sensing image registration and deep learning.\end{IEEEbiography}

\begin{IEEEbiography}[{\includegraphics[width=1in,height=1.25in,clip,keepaspectratio]{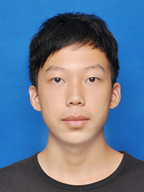}}]{Yi Dai}received the B.S. degree from Wuhan Polytechnic University, Wuhan, China, in 2024. He is currently pursuing the M.Eng. degree with the Remote Sensing Image Processing and Fusion Group, School of Electronic Engineering, Xidian University.\end{IEEEbiography}

\begin{IEEEbiography}[{\includegraphics[width=1in,height=1.25in,clip,keepaspectratio]{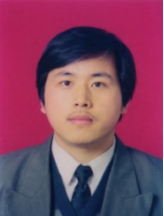}}]{Ming Li}(Member, IEEE) received the B.S. degree in electrical engineering and the M.S. and Ph.D. degrees in signal processing from Xidian University, Xi’an, China, in 1987, 1990, and 2007, respectively. 
	
In 1987, he joined the Department of Electronic Engineering, Xidian University. Currently, he is a Professor with the National Key Laboratory of Radar Signal Processing, Xidian University. His research interests include adaptive signal processing, detection theory, ultrawideband, and synthetic aperture radar (SAR) image processing.
\end{IEEEbiography}

\end{document}